\RequirePackage{fix-cm}
\documentclass[smallextended]{svjour3}       
\smartqed  
\usepackage{graphicx}
\usepackage{amsfonts}
\usepackage{kpfonts}
\usepackage{siunitx}
\usepackage{url}
\usepackage[draft]{fixme}
\newcommand{\etal}{\mbox{\emph{et al.\ }}}

\usepackage{dsfont}
\def\ONE#1{\mathds{1}\!\left\{#1\right\}}
\begin{document}

\title{Cell identification in whole-brain multiview images of neural activation
maps\thanks{MP was supported by a grant from Ente Cassa di Risparmio di Firenze.
LS and FSP were supported by the Italian National Flagship
NanoMAX and by the European Union Seventh Framework Program (FP7/2007-2013)
under Grant Agreements n. 604102 (Human Brain Project) and 284464 (LASERLAB-EUROPE).
We also wish to acknowledge a hardware grant from NVIDIA.}
}


\author{Marco Paciscopi  \and
        Ludovico Silvestri \and
        Francesco Saverio Pavone \and
        Paolo Frasconi
}


\institute{M. Paciscopi \at
              DINFO, Universit\`a di Firenze \\
              \email{marco.paciscopi@stud.unifi.it}           
           \and
           L. Silvestri \at
              Istituto Nazionale di Ottica (INO-CNR) \\
              LENS, Universit\`a di Firenze \\
              Tel.: +39-055-457-2504\\
              \email{silvestri@lens.unifi.it}           
           \and
           F. Pavone \at
              LENS and Dipartimento di Fisica e Astronomia, Universit\`a di Firenze \\
              Istituto Nazionale di Ottica (INO-CNR) \\
              Tel.: +39-055-457-2480\\
              \email{francesco.pavone@unifi.it}           
           \and
           P. Frasconi \at
              DINFO, Universit\`a di Firenze \\
              Tel.: +39-055-275-8647\\
              \email{paolo.frasconi@unifi.it}           
}

\date{Received: date / Accepted: date}

\maketitle

\begin{abstract}
  We present a scalable method for brain cell identification in
  multiview confocal light sheet microscopy images. Our algorithmic
  pipeline includes a hierarchical registration approach and a novel
  multiview version of semantic deconvolution that simultaneously
  enhance visibility of fluorescent cell bodies, equalize their
  contrast, and fuses adjacent views into a single 3D images on which
  cell identification is performed with mean shift.

  We present empirical results on a whole-brain image of an adult
  Arc-dVenus mouse acquired at $\SI{4}{\micro\meter}$
  resolution. Based on an annotated test volume containing 3278 cells,
  our algorithm achieves an $F_1$ measure of 0.89.

\end{abstract}

\section{Introduction}
\label{intro}
Understanding the basic principles of brain dynamics is one of the biggest challenges for contemporary
science. The presence of a complex network of short- and long-range connections between neurons
results into a tight functional coupling between different areas of the brain, since each single neural
cell can be elicited or inhibited by a cohort of neurons distributed throughout the whole encephalon.
Therefore, different brain states associated with specific behavioral or cognitive states result in
distinct patterns of neuronal activation \cite{alivisatos_brain_2012}. The technical ability to map such
patterns with single-cell resolution would provide a clearer view of brain activity and of its relation
with the underlying anatomical architecture \cite{oh_mesoscale_2014}.

State-of-the art techniques for \emph{in vivo} neuronal activity imaging are usually limited by coarse
resolution or restricted field of view. For instance, functional magnetic resonance imaging (fMRI) can
monitor neuronal activity in vivo throughout the whole brain, but with a spatial resolution too coarse
to distinguish single cells \cite{logothetis_what_2008}. On the other hand, electrophysiology recordings
or two-photon optical functional imaging allow inspecting neuronal activity with high resolution, but
only on a small area \cite{kerr_imaging_2008}. To overcome these limitations, and afford brain-wide
cellular-resolution neuronal activation maps, a complementary approach based on \emph{ex vivo} mapping
of immediate early genes (IEGs) expression has been proposed in the recent years
\cite{vousden_whole-brain_2014,kim_mapping_2015}. Indeed, several transgenic mouse strains have been developed
showing expression of a fluorescent protein under the promoter of one of the two main IEGs (Arc and
c-Fos), thus resulting in fluorescent tagging of activated neurons
\cite{barth_visualizing_2007,eguchi_vivo_2009,guenthner_permanent_2013}. Since endogenous fluorescence is
preserved after tissue fixation, high-throughput \emph{ex vivo} microscopy and image analysis can be
used to quantify expression of one IEG with cellular resolution across the whole mouse brain. Vousden
\cite{vousden_whole-brain_2014} and Kim \cite{kim_mapping_2015}, together with their co-workers, demonstrated
this approach by using Serial Two-Photon sectioning tomography (STP) \cite{ragan_serial_2012} and 2D
cell localization based either on 2D segmentation \cite{vousden_whole-brain_2014} or convolutional neuronal
networks \cite{kim_mapping_2015}. Nevertheless, since STP typically images only one optical section
($\approx \SI{1}{\micro\meter}$) every $50-\SI{100}{\micro\meter}$, only a small fraction of the
brain volume ($\approx 1-2$ \%) was actually sampled in these works.

Full volumetric imaging of macroscopic specimens with micrometric resolution is possible using light
sheet microscopy (LSM) coupled with tissue clearing
\cite{dodt_ultramicroscopy:_2007,silvestri_confocal_2012,costantini_versatile_2015}. However, residual
scattering of light and other artifacts due to imperfect sample clearing introduce quite large variability
of quality and contrast in LSM images, thus challenging state-of-the-art image analysis methods.
For example, cell detection procedures like
NeuroGPS~\cite{quan_neurogps:_2013} or
DeadEasy~\cite{forero_deadeasy_2009} are based on morphological
analysis which is preceded by binarization. Finding a correct
binarization threshold in LSM images is hard because even spatially
close structures may have significantly different intensities.
To tackle the quality variability
problem, we recently developed an image processing method (Semantic Deconvolution, SD) to enhance the
structures of interest (fluorescent cell bodies) and equalize their contrast across the entire image
\cite{frasconi_large-scale_2014}. SD is based on a supervised learning approach. Given
knowledge of true cell coordinates in a small training volume, a neural network is trained to convert the original images
into ``ideal images'' where cell bodies are represented as 3D truncated Gaussians, and
everything else is dark. Once SD has been applied to
the images, simple localization algorithms like mean shift can reach very high performance scores \cite{frasconi_large-scale_2014}.

\begin{figure*}
\centering
\includegraphics[width=\textwidth]{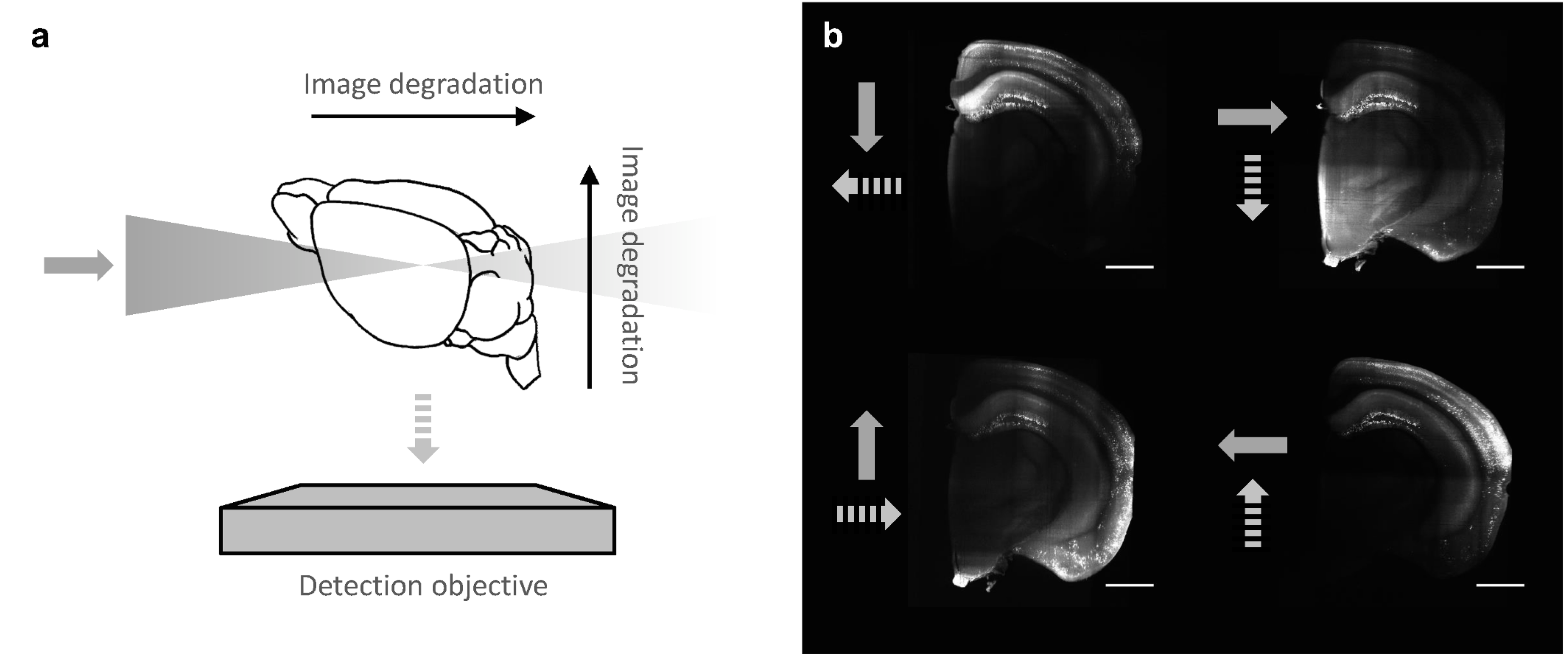}
\caption{Image degradation in LSM. (a) both excitation (solid arrow) and fluorescence light (dashed arrow)
are scattered and absorbed while traveling inside the specimen, leading to image degradation along two
orthogonal axes. (b) virtual coronal sections of half a mouse brain imaged with LSM at different angles.
The direction of the ingoing excitation light and of the outgoing fluorescence are depicted with solid and
dashed arrows, respectively. Scale bars, 1 mm.}
\label{fig:LSM}
\end{figure*}

When imaging entire or even half mouse brains, the effects of light scattering can be so strong that some
regions of the sample result almost completely dark when imaged with LSM (see Fig. \ref{fig:LSM}). Since
the spatial position of the ``dark'' and ``bright'' regions depends on the orientation of the sample with
respect to the excitation and detection optics (see Fig. \ref{fig:LSM}), the specimen can be rotated to
obtain a collection of images in which every part of the sample is in a ``bright'' region at least once.
Multiview LSM is indeed quite common in non-cleared specimens, like embryos
\cite{huisken_optical_2004,keller_reconstruction_2008}, and a number of algorithms for alignment and fusion
of multiple views have been developed in the last years
\cite{swoger_multi-view_2007,rubio-guivernau_wavelet-based_2012,preibisch_efficient_2014}. Several of these multiview alignment and
fusion methods rely on the presence of reference ``bright stars'' -- which in practice are fluorescent
nanospheres embedded in a gel surrounding the biological sample -- for highly precise alignment
\cite{rubio-guivernau_wavelet-based_2012,preibisch_efficient_2014}. Intensity--based registration algorithms
\cite{swoger_multi-view_2007} have less requirements in sample preparation, but can introduce alignment artifacts
in presence of slight specimen deformations between the acquisition of different views.

Here, we address the problem of extracting the localization maps of Arc-expressing neurons (tagged with
the fluorescent protein dVenus \cite{eguchi_vivo_2009}) in multiview LSM images of half a mouse brain.
Conventional multiview registration and fusion approaches -- which have been devised for samples which
are at least one order of magnitude smaller -- can be hardly applied with such a large sample: In fact,
on the one hand, sample embedding in a gel (which is mandatory for bead-based methods) is not compatible
with the clearing protocol (see~\S\ref{sec:sample}). On the other hand, intensity-based registration is
limited by the large variability of contrast in the same area for different views (see Fig. \ref{fig:LSM}).

\begin{figure*}
  \centering
  \includegraphics[width=\textwidth]{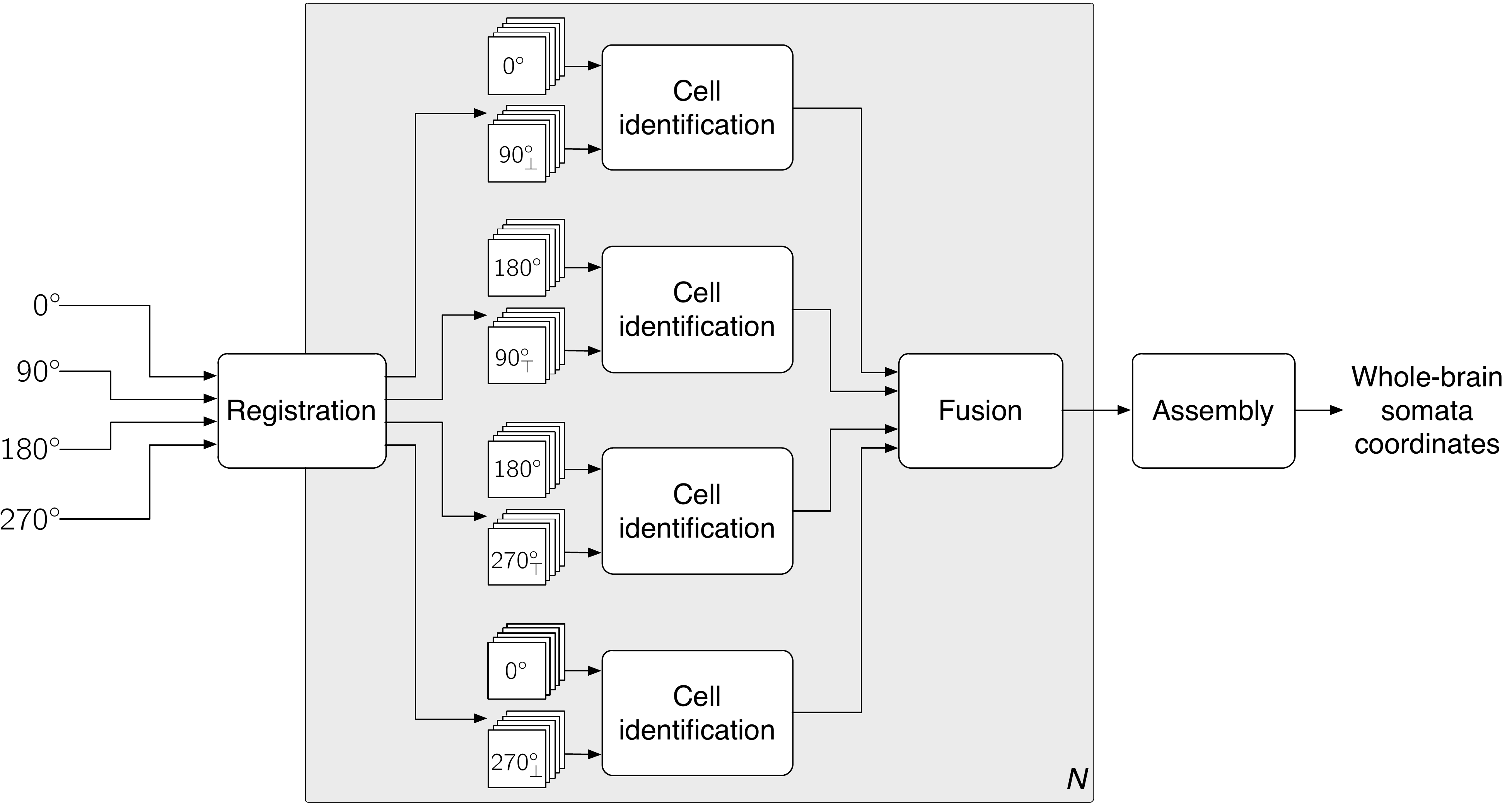}
  \caption{Overall processing pipeline for cell identification. $d^\circ_\perp$ and $d^\circ_\top$ (for $d=90,270$) denote
    substacks from the $d^\circ$ view aligned to the $0^\circ$ and the $180^\circ$ references, respectively.}
  \label{fig:pipeline}
\end{figure*}

We thus implemented a two-step registration approach consisting of a coarse global rigid registration, based
on manually-picked landmarks, followed by a local rigid alignment maximizing mutual information in pairs of substacks.
Pairs of adjacent views are then fused by extending the SD method to manage two images instead of one
as input. Afterwards, cell-localization is performed on the semantically deconvolved/fused images. Finally,
extracted cell positions from all the pairwise fused images are merged together in a single dataset
representing the full distribution of all Arc-expressing neurons in the sample. The entire pipeline is shown
in Figure~\ref{fig:pipeline}.

\section{Materials and Methods}

\subsection{Sample preparation}
\label{sec:sample}
Adult Arc-dVenus mice \cite{eguchi_vivo_2009} were fixed using standard transcardial perfusion with 4\%
paraformaldehyde (PFA) in phosphate buffered saline (PBS) solution. Brains were extracted from the skull,
post-fixed overnight in PFA 4\% @ ${4\;}^\circ$C, and stored in PBS 4\% @ ${4\;}^\circ$C. Our clearing protocol
is based on the one described by Becker et al. \cite{becker_chemical_2012}. Brains were first cut in two
halves along the longitudinal fissure, then dehydrated in a graded series of tetrhydrofuran (THF) in water
(50\%,70\%,80\%,90\%,96\%,100\% 1 h each, 100\% overnight). Afterwards, samples were cleared by immersion in
dibenzylether (DBE). DBE was changed three times (2h each) before imaging. Both THF and DBE were previously
filtered with aluminum oxide to remove fluorescence-quenching peroxides.

\subsection{Imaging}
\label{sec:imaging}
Samples have been imaged in a custom-made confocal light sheet microscope (CLSM) described in detail in
\cite{silvestri_confocal_2012}. Briefly, light from a $\SI{515}{\nano\meter}$ continuous wave laser is scanned by a galvanometric
mirror and slightly focused to produce a sheed of light in the sample, following the principle of digital
scanned laser light sheet microscopy (DLSM) \cite{keller_reconstruction_2008}. The light sheet produced lies at
the focal plane of a low-magnification detection objective (Leica HI PLAN 4X, numerical aperture 0.10), which
collects emitted fluorescent light. A de-scanning imaging system in the detection optical path creates a fixed
image of the scanning excitation laser line; at the position of this image a linear spatial filter (slit) is
placed to remove out-of-focus and scattered light. A third scanning mirror, inserted in a further imaging lens
system, reconstructs a bi-dimensional image on the chip of a high-sensitivity camera. A fluorescence filter
placed inside this third scanning system blocks all stray excitation light, allowing only fluorescence emission
to be collected by the camera. The sample chamber is mounted on a motorized system allowing specimen motion
along 3 perpendicular axis and rotation along the axis perpendicular to both the excitation and the detection directions.

The transfer function of the microscope (Point Spread Function, PSF) is quite anisotropic as its width
along the detection axis is 3-4 times larger than on the illumination plane \cite{silvestri_confocal_2012}. Therefore,
cell bodies usually appear as prolate ellipsoids rather than spheres; the orientation of the longer axis of the ellipsoid
depends on the angle from where the volume is imaged (see Supplementary Figure 1).
Although the resolution is anisotropic, we chose to use an isotropic sampling volume (voxel) of $\SI{4}{\micro\meter}$ side.
This choice simplifies multiview fusion as no resampling of the data is needed.
Several parallel stacks were
collected to cover the entire sample volume, since the field of view of the camera resulted in
$\approx 2 \times \SI{2}{\square\milli\meter}$. A small overlap of about $\SI{200}{\micro\meter}$ was introduced
between adjacent image stacks to allow subsequent stitching using the TeraStitcher software
\cite{bria_terastitcher-tool_2011}. Samples were imaged from 4 different angles separated by $90^\circ$.



\subsection{Multiview Coarse-to-Fine Registration}
\label{sec:registration}

There are several brain regions which are well visible in one view
only, and several other regions where contrast is not equally high in
all views. Thus, registering the images and exploiting the
complementary information in the four views can be expected to improve
cell identification accuracy.

In spite of the fact that the four views were obtained by simply
rotating the specimen at four different angles, a rigid transformation
is not sufficient to obtain a good registration. This is mainly due to
the incremental errors introduced during the initial stitching of the
CLSM tiles (the TeraStitcher software~\cite{bria_terastitcher-tool_2011} was
used for this purpose) which produce slight but noticeable non-linear
deformations of the images. The solution suggested in the present work
is a simple hierarchical algorithm that starts from a coarse rigid
transformation and then performs local refinements to reduce
the misalignments due to non-rigid deformations.
\begin{figure*}
  \centering
  \includegraphics[width=\textwidth]{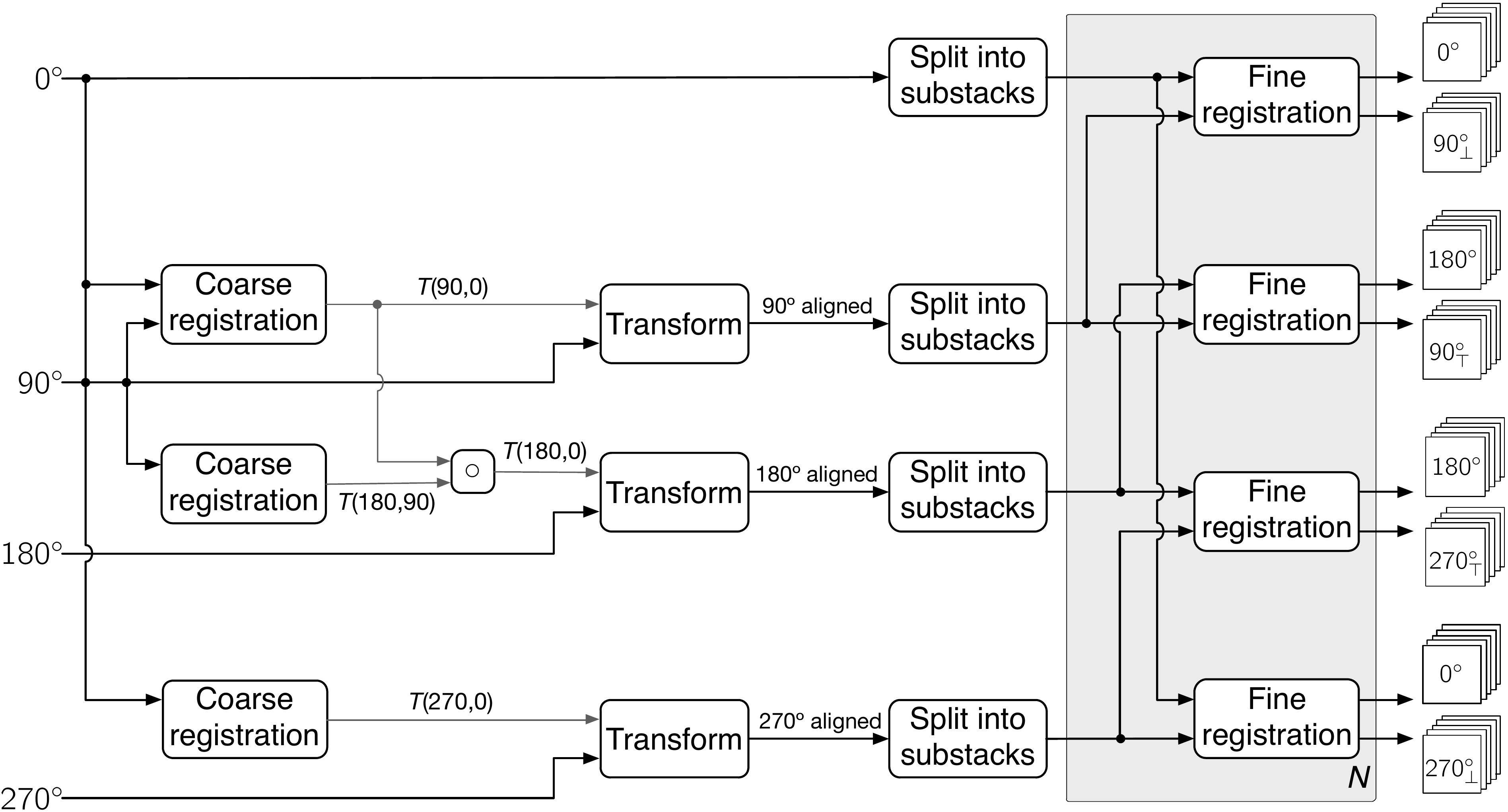}
  \caption{Registration pipeline. The inputs are four stitched volumes
    from orthogonal views.  $T(a,b)$ are the coarse transformations
    mapping from view $a$ to view $b$.  The shaded box is executed in
    parallel for all substacks. At the end, four paired sets of
    substacks associated with adjacent views are produced.}
  \label{fig:pipeline-registration}
\end{figure*}
The hierarchical registration pipeline is shown
in Figure~\ref{fig:pipeline-registration}.
The coarse registration stage is run three times in order to align the
target views $90^\circ$, $180^\circ$, and $270^\circ$ to the reference
$0^\circ$ view. As detailed below, coarse registration is based on a
small set of manually annotated fiducial points. The registrations
$90^\circ$ to $0^\circ$ and $270^\circ$ to $0^\circ$ can be performed
directly. However, opposite views $0^\circ$ and $180^\circ$ share an
insufficient portion of visible brain image to allow the extraction of
fiducial points. Hence, we first registered $180^\circ$ to $90^\circ$,
obtaining a transformation $T(180,90)$, and then computed the
transformation $T(180,0)$ as the composed transformation
$T(90,0) \circ T(180,90)$. 

In the middle of the registration pipeline, the four 3D volumes are
split into $N$ overlapping substacks as
in~\cite{frasconi_large-scale_2014} and subsequent operations (shaded
area in Figure~\ref{fig:pipeline-registration}) are performed
independently on adjacent pairs of substacks. Splitting large volumes
into smaller portions has several advantages:
\begin{itemize}
\item the transformation required for image registration is
  approximately rigid at the local level (see
  \S\ref{sec:registration});
\item operations on substacks can be carried out in parallel on a
  computing cluster;
\item the regional illumination variability can be factored out by
  performing thresholding at the substack level
  (see~\S\ref{sec:somata-identification});
\item the running time of the mean shift algorithm used for somata
  identification (see~\S\ref{sec:somata-identification}) is
  significantly reduced by operating at the local level.
\end{itemize}

The fine registration stage takes substacks from the $0^\circ$ and the
$180^\circ$ views as the reference images and substacks from the
$90^\circ$ and the $270^\circ$ views as the target images (i.e., it is
run four times in total). Fine registration between opposite views is
not run during this processing stage. This is for two reasons. First,
for most substacks too few cells are visible in both opposite views. Second, as
explained in~\S~\ref{sec:imaging}, spheres are deformed into prolate
ellipsoids whose longer axis orientation is the same for opposite
views ($0^\circ$-$180^\circ$ and $90^\circ$-$270^\circ$), where the
detection axis lies along the same line (see Supplementary Figure~1).
Our scheme ensures that any pair of aligned substacks has a consistent
orientation of deformation and this is useful to help generalization
of the neural network in the subsequent semantic deconvolution step.

The fine registration module is preceded by a test checking for black
regions\footnote{In practice we discard a substack view if the
  high foreground level (defined by the $\theta_2$ threshold explained in
  \S\ref{sec:somata-identification}) is below 30.}. If both views are
black the pipeline is terminated returning an empty list for that pair
of substack views.
In the following we provide additional details on the coarse and fine registration modules.

\paragraph{Coarse Registration.}
In this step, we estimate the 3D rigid transformation between pairs of
views of whole brain images. For this purpose, we manually annotated
the four views with 15 corresponding markers using the Vaa3D
software~\cite{peng_v3d_2010}. While in principle just three
correspondences are sufficient to estimate a 3D rigid transformation,
a greater number is required to compensate the unavoidable
imprecisions in the human-created landmarks due to low resolution, high
anisotropy, and illumination differences.
In order to improve the robustness of the solution, we applied the
RANSAC outlier rejection procedure
\cite{fischler_random_1981} to the list of landmark
correspondences. 
%
%
%
The resulting absolute orientation problem was solved using the
Arun \etal method \cite{arun_least-squares_1987}, which uses a closed-form
optimization algorithm and 3D points as registration features. This
technique solves a constrained least squares problem, based on the
computation of the Singular Value Decomposition (SVD) on a matrix
derived from the rotation component of the rigid transformation.
The estimated rigid transformation is finally applied to the whole 3D
volume.  Voxel intensities in the output image are assigned using the
nearest neighbor interpolation algorithm.

\paragraph{Fine Registration based on Mutual Information Metric.}
After coarse registration, a local fine-tuning procedure was employed to
improve the quality of the alignment. For this purpose, we split the
whole brain image into small substacks of size $91\times 90\times 90$ and registered
each pair of substacks by maximizing a mutual information metric
\cite{viola_alignment_1997}.
In particular, we used the approach proposed by Mattes \etal
\cite{mattes_nonrigid_2001} where the spatial samples used to
estimate the mutual information are retrieved at the beginning of the
process and remain unchanged during the optimization.
Assuming that the $g(x|\mu)$
is a rigid deformation from the coordinate frame of the test volume
($V_T$) to the reference domain ($V_R$), where $\mu$ is the set of
transformation parameters to be estimated, $f_T (g(x|\mu))$ represents
the transformed test volume voxel associated to the reference volume
voxel $f_R(x)$. To align $f_R(x)$ to the transformed test image
$f_T(x)$, the registration problem consists of determining the set of
parameters $\mu^{*}$ that minimize the negative mutual information $S$:
\begin{equation}
\mu^*=\mathrm{arg}\min_{\mu}S(f_R(x),f_T(g(x|\mu))).
\end{equation}
This step requires the estimation of the joint and
marginal intensity distributions of the reference and test images. Densities were estimated
from a representative sample of voxels of
both the images, using cubic B-spline Parzen
windows for smoothing. 
We determined that a sample of $10^5$ voxels (corresponding to 13 percent of the registered
volume) was sufficient. Using more samples would just increase running time without improving the
quality of registration.

\subsection{Ground truth}
\label{sec:ground-truth}
Knowledge of the true locations of somata centers is required for training the neural network
used in the semantic deconvolution step (see \S\ref{sec:semantic}) and for performance assessment (see
\S\ref{sec:performance-evaluation}).  For this purpose, we manually annotated a
random set of 56 substacks of size $91\times 90\times 90$ from
different anatomical regions in order to get a rich and diverse
collection of cases.

For each view of these substacks, we marked somata that were visible
in that view, using a modified version of the Vaa3D
program~\cite{peng_v3d_2010} that incorporates a three-dimensional local
cell detector based on mean-shift.
Landmarks from different views where subsequently merged using max-weighted bipartite
matching \cite{galil_efficient_1986}. For this purpose, we created
a weighted bipartite graph $G=(V,E)$, with a weight function
$w: E\rightarrow \mathbb{R}$ and bipartition $(V_1,V_2)$ where $V_1$
and $V_2$ correspond to manually annotated soma centers in the first
and second view, respectively. The weight of edge $(v_i,v_j)$, where
$v_i \in V_1$ and $v_j \in V_2$, was set to $w_{ij}=1/d_{ij}$ being
$d_{ij}$ the Euclidean distance between the landmarks
$v_i$ and $v_j$. The merged set of landmarks $M$ was then obtained as
follows:
\begin{itemize}
\item all unmatched vertices were added to $M$ (these are somata that
  are visible in one view only)
\item matched vertices correspond either to somata that are visible in
  both views, or to somata that are visible in one view but happen to
  be close in space; hence, if $d_{ij}\leq d^*$ we added to $M$ the
  middle point between $v_i$ and $v_j$ and if $d_{ij}> d^*$ we added
  the two landmarks separately; the threshold $d^*$ was set to $3$
  voxels which is slightly above the maximum soma radius at the image
  resolution.
\end{itemize}

\subsection{Content-Based Image Fusion}
\label{sec:fusion}
In previous work on LSM
imaging, it has been suggested to use content-based weighting in order
to fuse images taken from multiple angles into a single isotropic
volume \cite{preibisch_mosaicing_2008}. This method was applied to
\textit{whole} images following registration (global registration is
feasible in \cite{preibisch_mosaicing_2008} only thanks to the addition
of fluorescent beads to the rigid agarose medium) but it can also
be applied to the small substacks used in our setting.

Content-based fusion aims to minimize the blurred
parts of each single image, caused by artifacts of the microscopy, and
enhance the sharp ones. The adopted strategy consists of computing a
weighted average of voxel intensities of each view with their
corresponding entropy mask, estimated in the local neighborhood of
each voxel. Given two 3D aligned views of the same substack, $V_i$ and
$V_j$ , and their regional entropies $H_i(x,y,z)$ and $H_j(x,y,z)$,
the output fused tensor $V_{fused}$ is computed as follows:
\begin{equation}
V_{fused}(x,y,z)=\frac{100^{H_i(x,y,z)}V_i(x,y,z) + 100^{H_j(x,y,z)}V_j(x,y,z)}{100^{H_i(x,y,z)} + 100^{H_j(x,y,z)}}
\end{equation}
The entropy functions have been used as exponents to underweight the
entropy of all the blurred regions of the views that are not
completely uniform. The local entropy has been estimated in each voxel
from the intensity histogram retrieved by a window centered in that voxel
with a side length of $9$ pixels.

\subsection{Semantic deconvolution}
\label{sec:semantic}
Images acquired by CLSM generally suffer from significant contrast
variability, mainly due to inhomogeneous optical clearing and to
different depths that are traveled by the laser beam.  The problem is
exacerbated when using a high energy laser to penetrate thick tissues, since this can lead to voxel saturation close to
the laser entry point.
Semantic deconvolution (SD) has been shown to be a very effective preprocessing
step in the cell detection pipeline, selectively enhancing contrast
for the objects of interest and significantly boosting precision and
recall of cell detection~\cite{frasconi_large-scale_2014}.

SD is a supervised technique consisting of a neural network trained to
map native input images into ideal target images where only objects of
interest (somata in this case) are preserved and have uniform
visibility.  The algorithm for constructing the target image can be
summarized as follows. Given a set of soma centers
$\mathcal{D} = \left\{(x^{(i)}, y^{(i)}, z^{(i)}), i=1,\dots,n
\right\}$
(obtained by manual annotation), we first construct a 3D image with
intensity
$$
I(x,y,z) = \left\{
  \begin{array}{ll}
    1 & \mbox{if $(x,y,z) \in \mathcal{D}$}\\
    0 & \mbox{otherwise}
  \end{array}
\right.
$$
and then apply to $I$ a Gaussian filter with kernel standard
deviation $\sigma$, obtaining the target image $Y$. To ensure that
target blobs remain spatially well separated, the filter is truncated
at $2\sigma/3$.

The neural network is trained to map cubic patches of size $s^3$ of
the original image into corresponding cubic patches of the same size
in the target image. The network learns to enhance the visibility of
true soma and to reduce the visibility of voxels belonging to other fluorescently labeled
structures such as dendrites and axons. After training, the network is applied to the original 3D
image $X$ in a convolutional fashion in order to obtain the output
image $D$, i.e. for every tuple $(x,y,z)$ of voxel coordinates, we compute
\begin{equation}
  \label{eq:semantic}
  D(x,y,z) = \frac{1}{(2s+1)^3}
  \sum_{i,j,k=-s}^s
  F_{x-i,y-j,z-k}(i+1-x,j+1-y,k+1-z)
\end{equation}
where $F_{a,b,c}$ denotes the 3D tensor at the output of the neural
network when its input consists of the input patch
$X(a-s:a+s,b-s:b+s,c-s:c+s)$.

\begin{figure*}
  \centering
  \includegraphics[width=\textwidth]{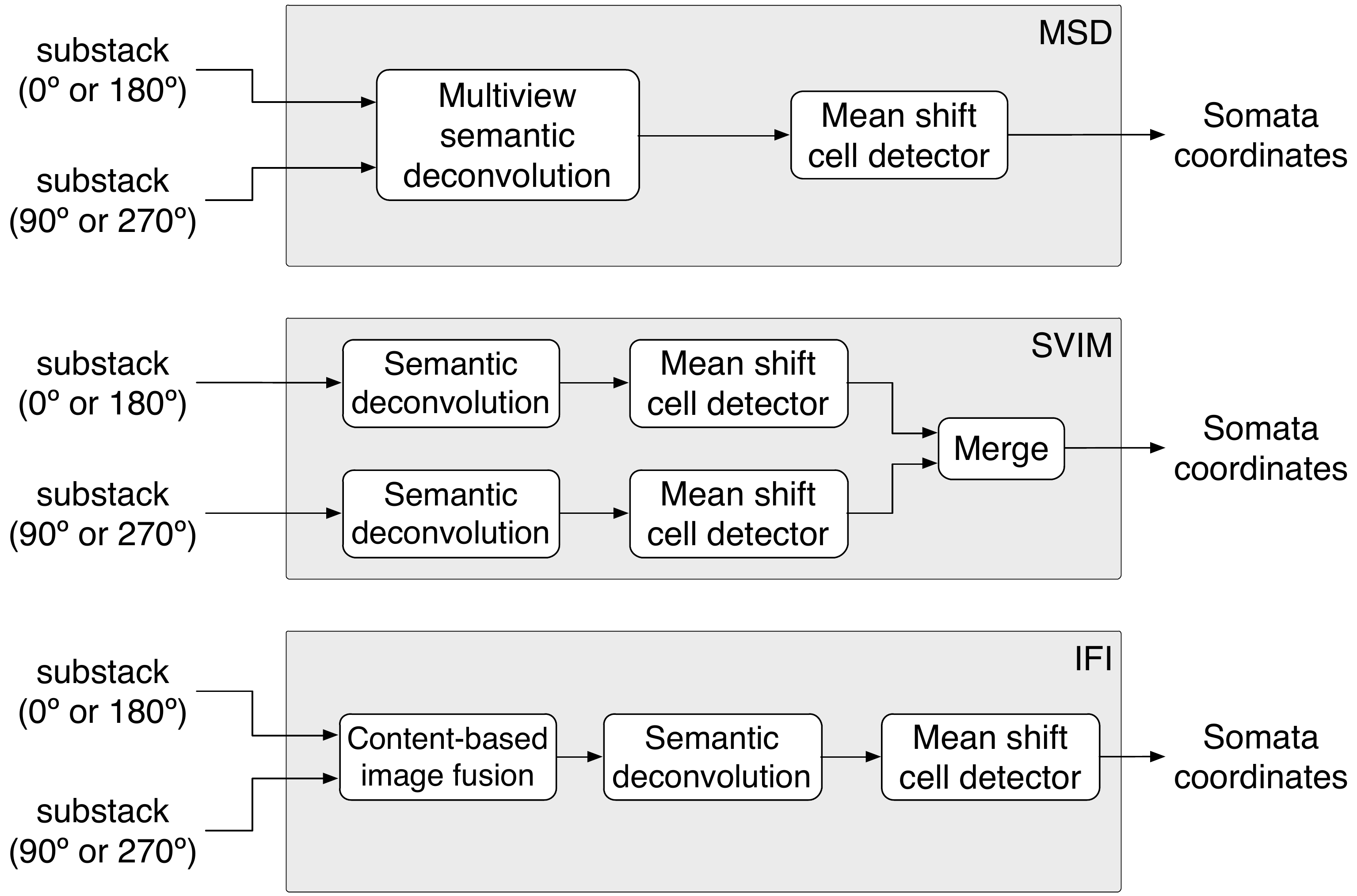}
  \caption{Three alternative options for multiview cell detection}
  \label{fig:pipelines-identification}
\end{figure*}

\subsection{Multiview cell detection}
\label{sec:multiview-semantic}
Dealing with multiview images offers at least three different options
for setting up the cell identification pipeline (see
Figure~\ref{fig:pipelines-identification}):
\begin{enumerate}
\item \textit{Single view identification and merge} (SVIM): Identify
  cells separately in each view and subsequently merge the sets of
  cells;
\item \textit{Identification on fused image} (IFI): Fuse views after
  registration into a single image and perform cell identification on
  the merged image;
\item \textit{Multiview semantic deconvolution} (MSD): Develop a
  specialized SD technique that simultaneously (1)
  merges different views into a single images and (2) performs
  selective contrast enhancement;
\end{enumerate}
The pipeline of~\cite{frasconi_large-scale_2014} (SD followed by
mean shift) can be applied immediately to the first two options. One
of the contributions of this paper is the development of a novel
multiview SD module for enabling the third
option. Our results (see Section~\ref{sec:MSD-is-better}) show that
MSD outperforms both SVIM and IFI.

\begin{figure*}
  \centering
  \includegraphics[width=\textwidth]{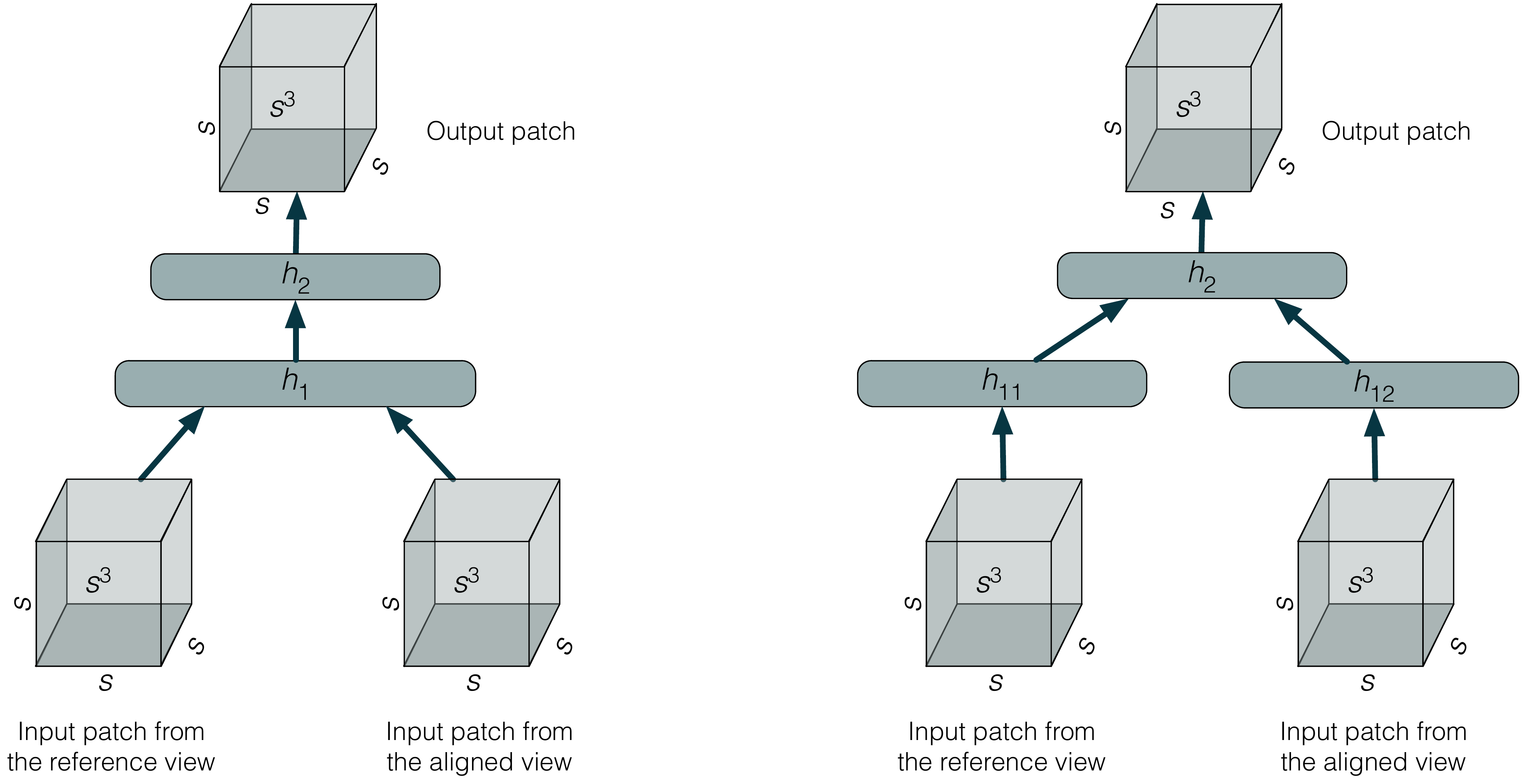}
  \caption{Neural network architectures for multiview semantic deconvolution: flat (left) and columnar (right).}
  \label{fig:MSD-architectures}
\end{figure*}

We investigated two alternative neural network architectures for MSD,
as shown in Figure~\ref{fig:MSD-architectures}. In both cases the
network takes as input a pair of cubic patches of size
$s\times s\times s$. The first patch is taken from a reference substack (either $0^\circ$ or $180^\circ$) and the second patch is
taken from the corresponding registered substack (either $90^\circ$ or
$270^\circ$) following affine transformation. Both patches are taken
from the same local coordinates.  The network is trained to
predict as output the cubic patch of size $s\times s\times s$ in the
target image, at the same local coordinates as the input patches. Each
output voxel intensity is regarded as the probability that a neural
soma occurs at that position. Thus each of the $s^3$ outputs of the
network has a logistic activation function and the log-loss (or
negative cross entropy) is minimized during training:
\begin{equation}
  \label{eq:loss}
  E(W) = \sum_{i} \sum_{j} \sum_{k} y(i,j,k) \log F(i,j,k;W) + (1-y(i,j,k)) \log (1-F(i,j,k;W))
\end{equation}
where $W$ are the neural network weights, $y$ denotes the target patch, $F$ the patch generated by the neural network,
$i$ ranges over the
training set substacks, $j$ over the patches in
substack $i$, and $k$ over the $s^3$ voxels of the patch.

\begin{figure*}
  \centering
  \includegraphics[width=\textwidth]{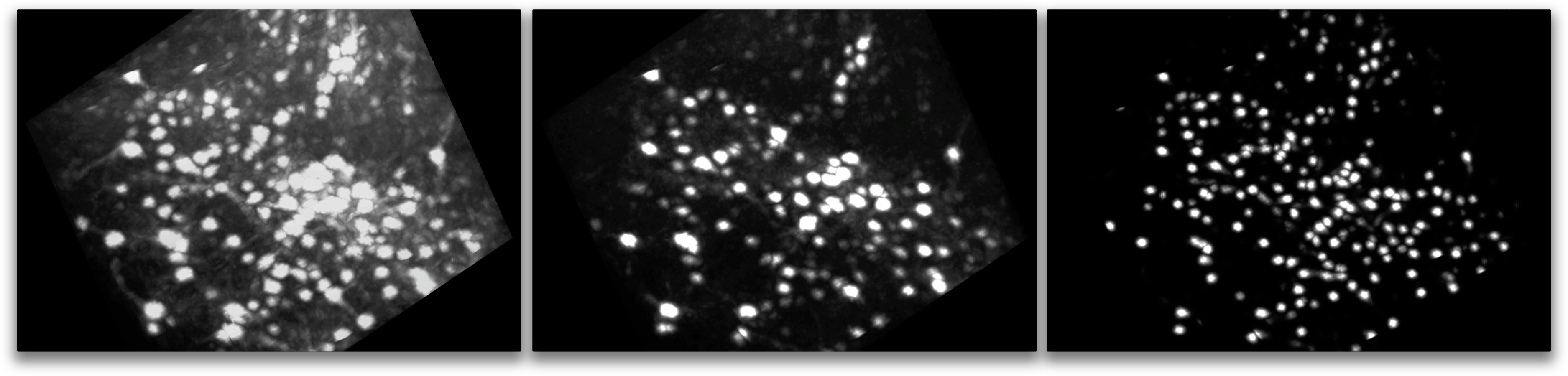}
  \caption{Illustration of multiview semantic deconvolution. Left:
    $0^\circ$ view; middle: $90^\circ$ view; right: output image
    produced by semantic deconvolution. Best viewed by zooming in on the electronic version.}
  \label{fig:sd}
\end{figure*}

We also experimented alternative ways of training the network. The
first procedure uses greedy layer-wise pretraining, using unsupervised
learning of restricted Boltzmann machines as originally introduced
in~\cite{hinton_fast_2006}, followed by supervised backpropagation
fine-tuning. In this case, logistic units, $f(z)=1/(1+exp(-z))$, are
used in all hidden layers.  The second procedure does not use
pretraining and learns by backpropagation only, using rectified linear
units, $f(z) = \max(0,z)$, in all hidden layers. We also experimented with a ``masked training'' procedure
(loosely reminiscent of dropout~\cite{baldi_dropout_2014}) where one of the two views is randomly
zeroed completely, in order to help the network to produce visible
somata in the output image even when they are visible in one of the
two views only.

\subsection{Somata identification}
\label{sec:somata-identification}
We formulate the task of somata identification in terms of clustering
together voxels belonging to the same soma. This is performed in three
steps:
\begin{description}
\item[Thresholding.] Voxels with very low intensity are unlikely to be found
  inside a soma and are therefore discarded in this step. Thresholding
  offers two advantages: computational efficiency of the subsequent
  clustering step, and reduction of false positives. Because of the
  high contrast and illumination variability in different image
  regions, any global thresholding technique is prone to severe errors
  (such as suppression of somata in low-visibility regions, if the
  intensity threshold is too high, or production of too many non-soma
  clusters, if the intensity threshold is too low). Local thresholding
  algorithms, on the other hand, are computationally costly,
  especially for large 3D images.  As a reasonable compromise, we
  operate at the level of substacks and apply a multi-threshold
  algorithm~\cite{sahoo_survey_1988} based on maximum
  entropy~\cite{kapur_new_1985}. In this way we obtain two
  substack-local thresholds $\theta_1$ and $\theta_2$. Voxels with
  intensity below $\theta_1$ are then discarded, yielding a set
  of foreground voxels $L$.

\item[Seeding.] The clustering procedure starts from a set of
  carefully selected voxels (seeds) which are good candidate soma
  centers. In order to be chosen as a seed, a point must satisfy two
  conditions: it must be a local maximum of the intensity, and must be contained in
  a region of sufficiently high intensity. Formally, the seeds set is
  defined as
  $$
  S = \{\vec{p}\in L: m(\vec{p})\} \cap \{\vec{p}\in L: \overline{I}(\vec{p},r) > \theta_1\}
  $$
  where $m(\vec{p})$ is true if $\vec{p}$ is a local maximum of the
  image intensity and $\overline{I}(\vec{p},r)$ is the average local
  intensity of the image in the $r$-neighborhood of $\vec{p}$, which
  we define in two alternative ways:
  \begin{equation}
    \label{eq:neighborhood}
    \overline{I}(\vec{p},r) = \left\{
    \begin{array}{lll}
      \displaystyle
      \frac{\sum_{\vec{q}} I(\vec{q})\ONE{\|\vec{p}-\vec{q}\|<r}}{\sum_{\vec{q}} \ONE{\|\vec{p}-\vec{q}\|<r}}& ~ & \mbox{(hard criterion)}\\
      ~\\
      \displaystyle
      \frac{\sum_{\vec{q}} I(\vec{q})\exp(-\|\vec{p}-\vec{q}\|/r)} {\sum_{\vec{q}} \exp(-\|\vec{p}-\vec{q}\|/r)}& ~ & \mbox{(soft criterion)}\\
    \end{array}
    \right.
  \end{equation}

  In both cases, $r$ controls the trade-off between false positives
  and false negatives in the identification procedure defined in the
  next step: large values reduce the number of seeds, thus decreasing
  false positives, while small values yield many seeds, decreasing
  false negatives. In practice we found that the optimal tradeoff
  occurs when $r$ is close to the expected radius of the fluorescent
  cell bodies and that the soft criterion yields better results.
\item[Mean shift clustering.] Starting from all elements of $S$, the
  mean shift algorithm~\cite{comaniciu_mean_2002} iterates until
  convergence the following two steps:
  \begin{enumerate}
  \item
    $$
    \vec{q}^{(i)} \leftarrow \frac{\sum_{\vec{p}\in L} I(\vec{p}) k(\vec{p},\vec{p}^{(i)};b) \vec{p}}{\sum_{\vec{p}\in L} I(\vec{p})k(\vec{p},\vec{p}^{(i)};b)}
    $$ 
  \item $\vec{p}^{(i)} \leftarrow \vec{q}^{(i)}$
  \end{enumerate}
  where $I(\vec{p})$ is the intensity of voxel $\vec{p}$ and
  $k(\vec{a},\vec{b};b)$ is a radial kernel function, parameterized by
  bandwidth $b$. If $k(\vec{a},\vec{b};b)$ is monotonically non
  increasing with $\|\vec{a}-\vec{b}\|$, the algorithm is guaranteed
  to converge. In practice we choose
  \begin{equation}
    \label{eq:kernel}
  k(\vec{a},\vec{b};b) = \left\{
    \begin{array}{ll}
      1 & \mbox{if $\|\vec{a}-\vec{b}\| \leq b$}\\
      0 & \mbox{otherwise}
    \end{array}
    \right.
  \end{equation}
  so that every mean corresponds to the ``baricenter'' (using
  intensities as masses) of the spherical image patch defined by the
  kernel.  At the end, duplicates are removed resulting in a set
  $\{\vec{p}^{(i)}\}$ of predicted somata centers.  Mean shift has
  been applied before to cell detection in zebrafish brain but in a
  very different setting, working in color space and with the purpose
  of filtering out false positive
  detections~\cite{liu_automated_2008}. In our approach mean shift
  works in coordinate space and is directly responsible for the
  identification of soma centers.
\end{description}

\subsection{Fusion and whole-brain assembly of cell coordinates}
\label{sec:fusion-assembly}
The identification module described in \S\ref{sec:somata-identification}
returns a list of cell body coordinates associated to every pair of adjacent views (i.e. four lists for each substack).
When a cell is visible in several views, its coordinates
may appear in more than one of these lists. The fusion module in
Figure~\ref{fig:pipeline} is in charge of removing duplicates and
merging results. Duplicates were removed by a merge approach based on
the Iterative Closest Point procedure (ICP)~\cite{Simon:1996:FAS:927342}. In order to give greater importance to
the pairs with the smallest distance, the rigid motion estimation was
accomplished solving a weighted least squares problem with a
quaternion-based method (Horn
\etal~\cite{Horn88closed-formsolution}). After aligning the point
clouds, all the computed correspondences were replaced by their
midpoint, while all the remaining points (those present in only one
list) were preserved.

Concerning the fusion procedure, there are five possible cases to
consider, depending on the number of non-empty lists (empty lists are also
returned if both views in a substack pair are black). The case are
detailed below:
\begin{enumerate}
\item All four lists are empty (trivial): return the empty list.
\item Only one list is non-empty (trivial): return that list.
\item If two lists are non-empty there are two sub-cases:
  \begin{enumerate}
  \item if the non-empty lists come from pairs having the same
    reference view (two possibilities: $(0^\circ,90^\circ_\perp)$ and
    $(0^\circ,270^\circ_\perp)$, or $(180^\circ,90^\circ_\top)$ and
    $(180^\circ,270^\circ_\top)$) it is sufficient to merge the
    lists;
  \item if they came from pairs having different references (four
    possibilities) then it means that some cells are detected in the
    $0^\circ$ and in the $180^\circ$ views. In this scenario, we
    attempt to compute the transformation from $180^\circ$ to
    $0^\circ$ using the fine registration procedure of
    \S~\ref{sec:registration}. Then, we apply the transformation to the cell coordinates and merge.
  \end{enumerate}
\item If there is only one empty list, then two lists comes from the
  same reference and two lists come from different references. We
  merge the first two lists, then compute the transformation
  $180^\circ$ to $0^\circ$ as above, and then merge the results.
\item If all lists are non-empty then we first merge the lists sharing
  the same reference, then apply the $180^\circ$ to $0^\circ$
  transformation, and finally merge the results.
\end{enumerate}
Once a unique (possibly empty) list is obtained for each substack, we
translate local coordinates into world coordinates in order to assemble the
whole-brain list of cell bodies.

\section{Results}

The images acquired with the method described in~\ref{sec:imaging} had
size $1823\times 1351 \times 2697$ and were split into 15,552 substacks
of size $91\times 90\times 90$ with an overlap of 16 voxels.  A random
subset of 10 manually annotated substacks (see \S\ref{sec:ground-truth})
was used for training the SD module. The remaining 45 substacks were
used to measure the performance of the overall cell identification
pipeline.

Unlike other typical machine learning settings, we deliberately used a
small volume for training the SD module since somata labeling requires
human intervention. Labeling 10 substacks required about 6 hours.
This effort is acceptable in relation to the processing time (see~\S\ref{sec:mapping}) and to the overall time required to prepare
the sample and to carry out imaging. Additionally, we expect that
labeling time can be amortized when working on several specimen
acquired with the same procedure.

\begin{figure*}
  \centering
  \includegraphics[width=\textwidth]{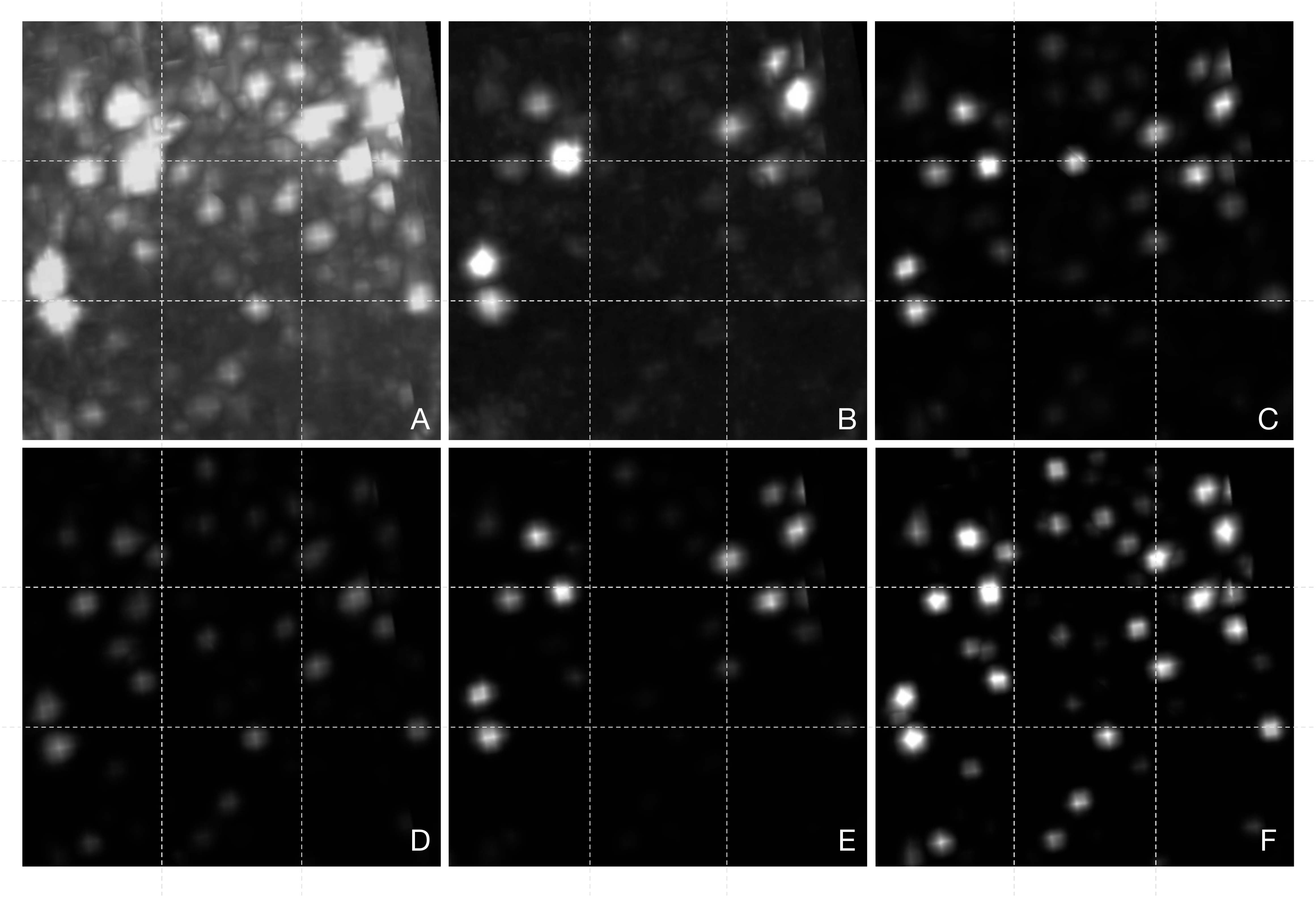}
  \caption{A: original $0^\circ$ view. B: original $90^\circ$ view. C:
    content-based image fusion of the two views followed by SD. D: SD on the $0^\circ$ view. E:
    SD on the $90^\circ$ view. F: multiview SD.}
  \label{fig:compare}
\end{figure*}

\begin{figure*}
  \centering
  \includegraphics[width=\textwidth]{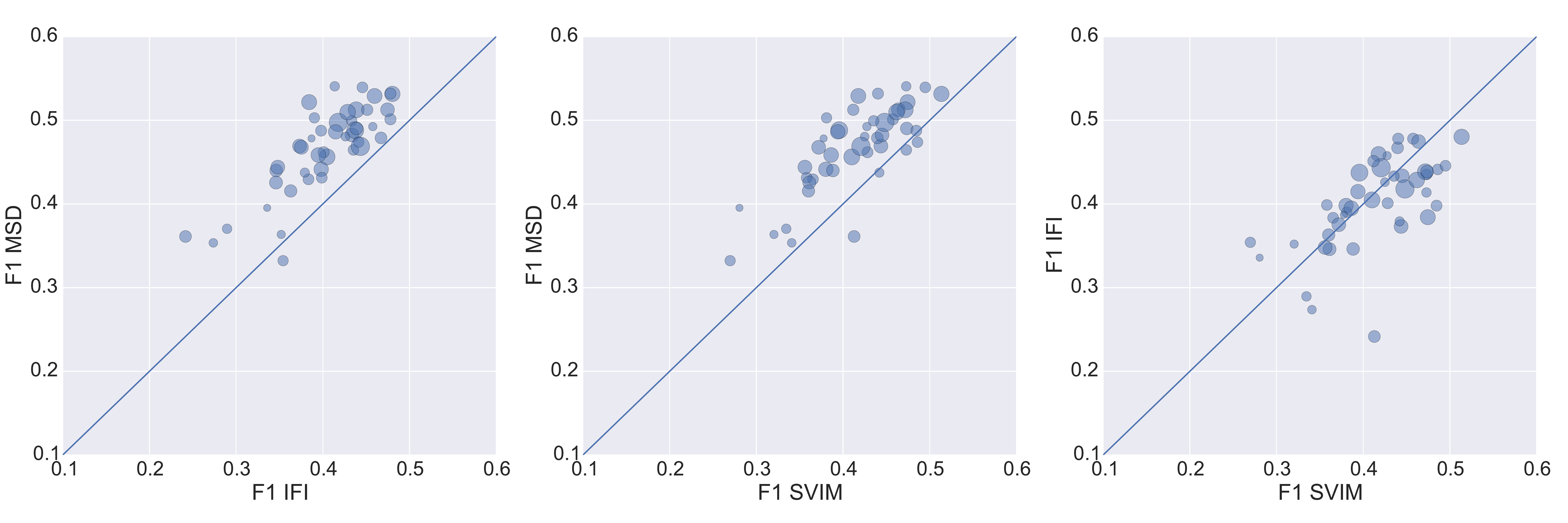}
  \caption{Voxel-level $F_1$ measure of semantically deconvolved images.
    Each dot is one of the 45 test substacks. Dot size is proportional to the
    number of somata in the substack.}
  \label{fig:f1-by-substack}
\end{figure*}

\subsection{Target images are better approximated by MSD}
\label{sec:MSD-is-better}
In this subsection, we compare the three alternative pipelines of
Figure~\ref{fig:pipelines-identification} \textit{before} the cell
identification stage. Unlike SVIM and IFI, multiview SD is trained to
capture the correlation between 3D patches in the two views. We found
that this approach enables a better approximation to the ideal output
image.  Figure~\ref{fig:compare} illustrates the differences among SD
applied to individual views (as in SVIM), SD on the fused image (as in
IFI), and MSD. Visual inspection suggests that some somata are
difficult to reconstruct from a single view or from the fused image
(this may lead to false negatives at the end of the identification
pipeline). In order to quantitatively compare the three approaches, we
binarized the images after SD using the maximum entropy threshold
$\theta_1$ as described in~\ref{sec:somata-identification} and
considered the binary classification problem where voxels belong to
the negative class iff they are dark in the ideal target
image. Since the vast majority of voxels are dark (negative), a very
high accuracy (defined as the fraction of correctly predicted voxels)
does not necessarily reflect a faithful reconstruction of the target
ideal image. For this reason, we consider the 
$F_1$ measure between the binarized target and semantically
deconvolved images. $F_1$ equals 1 for a perfect classifier.  For the
SVIM setting, we used the logical ``or'' between the binarized
outputs of SD from the two views. Results are shown in
Figure~\ref{fig:f1-by-substack}, (each dot corresponding to one of the
45 test substacks).  While there is no clear winner between IFI and
SVIM (p-value of 0.2 from the Wilcoxon signed-rank test),
MSD is better than IFI and SVIM on most substacks (p-values are both
below $10^{-7}$).

Low $F_1$ values indicate that the image fed to the subsequent cell
detection stage is likely harder to interpret but $F_1$ measures at the
level of voxels and at the level of cell detection do not necessarily
correlate perfectly: indeed, the voxel-level $F_1$ measure does not take
into account the different sizes of the visible cell bodies and the
spatial distribution of incorrect voxels. Performance of cell
detection is detailed in the remainder of this section.

\subsection{Measuring the performance of cell detection}
\label{sec:performance-evaluation}

For a given substack, we denote by $T$ the set of true somata centers
(according to the ground truth) and by $P$ the set of predicted
centers. We first match predictions to true centers by creating a
weighted bipartite graph $(T\cup P, T\times P)$ with edge weights
$w(\vec{t},\vec{p}) = 1/\|\vec{t}-\vec{p}\|$ for all $\vec{t}\in T$ and $\vec{p}\in P$. We
then run the max-weighted bipartite matching
algorithm~\cite{galil_efficient_1986} to obtain a set of matches
$M\subset T\times P$. We finally discard any match $(\vec{t},\vec{p})$ if
$\|\vec{t}-\vec{p}\|\geq 3.5$ voxels.
Unmatched true centers are counted as false negatives (FN), unmatched
predicted centers are counted as false positives (FP), and all matches
are counted as true positives (TP). We finally report precision
$P=\frac{\mathrm{TP}}{\mathrm{TP}+\mathrm{FP}}$, recall
$R=\frac{\mathrm{TP}}{\mathrm{TP}+\mathrm{FN}}$, and $F_1$ measure
$\frac{2PR}{P+R}$.

\subsection{Stability with respect to mean shift parameters}
\label{sec:stability}
One problem faced when designing algorithms for cell detection is that
performance may significantly depend on the various parameters of the
algorithm. If the optimal value of the parameters needs to be adjusted
locally in each image region, the algorithm cannot be easily applied at
large scale (this is particularly true for CLSM images because of
their inherent contrast variability). Ideally the algorithm should be
stable, i.e. (1) performance should not change significantly for small
perturbations of the parameters, and (2) the optimal value of
parameters should not change significantly in different regions of the
image.  Our cell identification algorithm depends on the seed ball
radius, $r$, and the mean shift bandwidth, $b$. We expect that
semantic deconvolution will increase the stability of mean shift with
respect to these parameters. In order to verify this speculation, we
setup the following experiment, using the SVIM and IFI pipelines of
Figure~\ref{fig:pipelines-identification}, first keeping the SD
modules and then suppressing them (i.e. using raw
images)\footnote{suppressing semantic deconvolution in the MSD
  pipeline would yield two raw images as in SVIM.}. In each of the four resulting
settings we studied stability as follows.  First, we computed the
parameters $r^*$ and $b^*$ that maximize the overall $F_1$ measure on
the whole available ground truth (45 labeled substacks). Then, for
each substack $S_i$, we computed the parameters $r^*_i$ and $b^*_i$
that locally maximize the $F_1$ measure on $S_i$. 

Results are reported
in Table~\ref{tab:stability} where $F_1^{\mathrm{glob}}$ denotes the
performance obtained by using the same globally optimal parameters in
every substack and $F_1^{\mathrm{loc}}$ the performance obtained by
using locally optimal parameters. 95\% confidence intervals were
computed by Monte Carlo simulation, taking 10000 samples from the
$F_1$ distribution defined by the probabilistic model described
in~\cite{goutte_probabilistic_2005}.
\newcommand\ci[3]{#1 ~[#2,#3]}
\begin{table*}
  \centering
  \begin{tabular}{l c c c c}
    ~  &\multicolumn{2}{c}{Raw}& \multicolumn{2}{c}{SD} \\
    ~ & $F_1^{\mathrm{glob}}$  & $F_1^{\mathrm{loc}}$ & $F_1^{\mathrm{glob}}$ & $F_1^{\mathrm{loc}}$ \\
    \hline\noalign{\smallskip}
    SVIM& 71.1 ~[69.8,72.4]    & 81.0 ~[80.0,82.0]    & 88.1 ~[87.3,88.9]    & 91.5 ~[90.8,92.2]\\
    IFI & 75.1 ~[73.9,76.2]    & 82.8 ~[81.8,83.8]    & 87.7 ~[86.8,88.5]    & 90.7 ~[89.9,91.4]\\
    MSD& -    & -    & 89.4 ~[88.6,90.2]    & 92.1 ~[91.4,92.7]\\
  \end{tabular}
  \caption{Stability of mean shift (95\% confidence intervals in brackets).}
  \label{tab:stability}
\end{table*}
Both SVIM and IFI result in a large difference between
$F_1^{\mathrm{glob}}$ and $F_1^{\mathrm{loc}}$ when running mean shift on raw
images. Differences are significantly reduced if mean shift is run
on semantically deconvolved images. Additionally, $F_1^{\mathrm{glob}}$
after SD is significantly better than $F_1^{\mathrm{loc}}$ on raw
images. These results show that SD improves performance significantly and stabilizes it with respect to
the parameter values.

\subsection{Comparing SVIM, IFI, and MSD}
\label{sec:performance}

\begin{figure*}
  \centering
  \includegraphics[width=\textwidth]{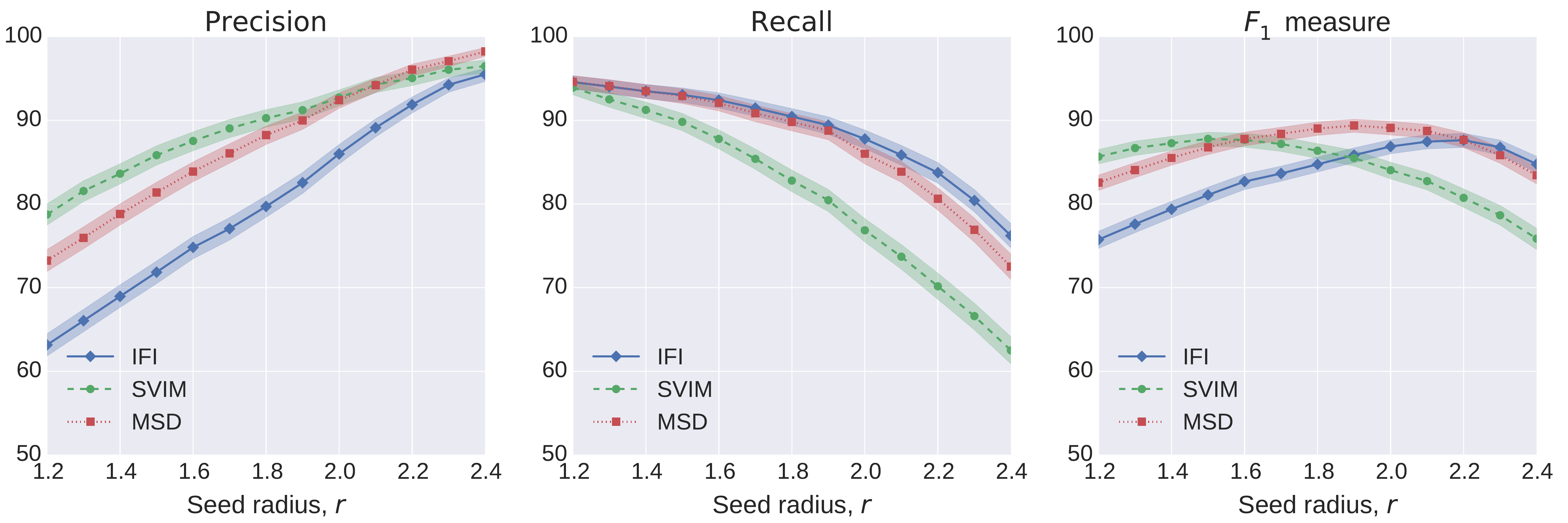}
  \caption{Performance of mean shift after IFI, SVIM, and MSD when changing the seed radius $r$.}
  \label{fig:perf-mean-shift}
\end{figure*}

Figure~\ref{fig:perf-mean-shift} reports precision, recall, and $F_1$
measure, when varying the radius of the seed ball, $r$, for the cell
identification algorithm.  As expected, $r$ controls the
tradeoff between false positives and false negatives and therefore
precision increases and recall decreases as the criterion for
accepting a local maximum as a seed for the mean shift algorithm
becomes more stringent. While the precision values of SVIM, IFI, and
MSD are relatively close for several $r$'s, recall is significantly
better for MSD. In particular, the improvement over SVIM means that
the neural network is capable of capturing the correlation between
views: learning to combine adjacent views into a single ideal image is
more effective than trying to recover two separate ideal images
independently from individual views.  Interestingly, the optimal
tradeoff (best
$F_1$ measure) is met when $r$ is close to the expected radius of
the visible fluorescent cell body (between 1 and 2 voxels in our data).
The optimal radius is slightly larger (1.9) in the case of
images processed by MSD since MSD tend to produce output images where
somata are brighter and larger compared to IFI and SVIM (see
Figures~\ref{fig:sd} and~\ref{fig:compare}).


\begin{figure*}
  \centering
  \includegraphics[width=\textwidth]{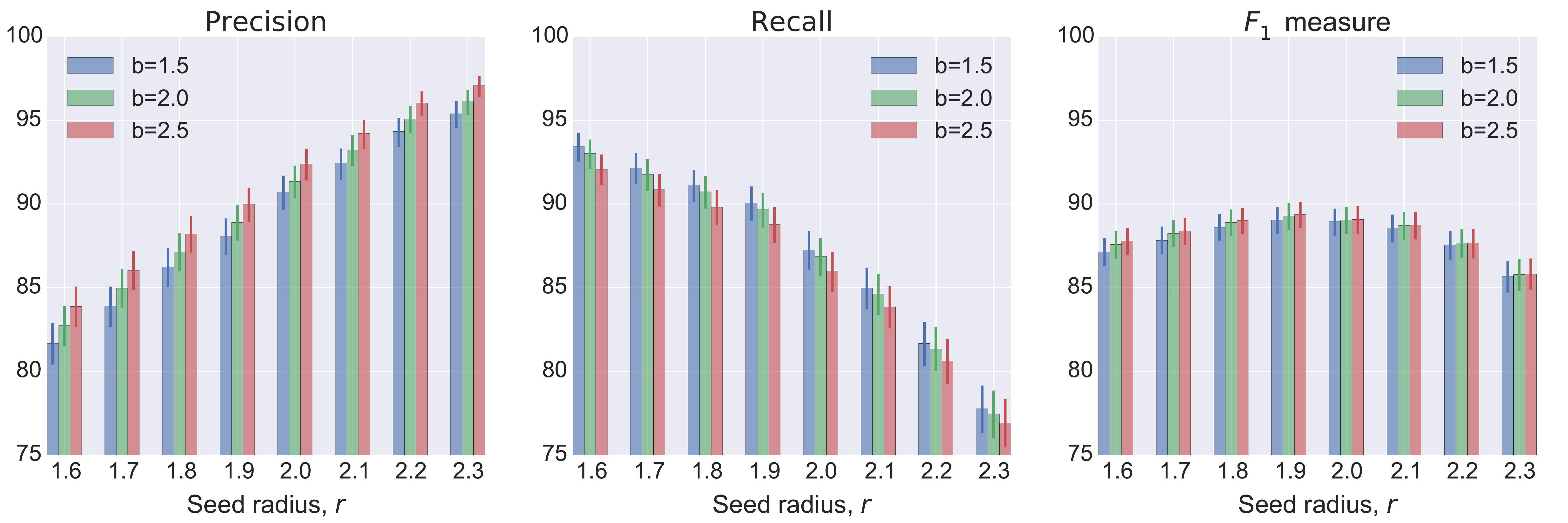}
  \caption{Performance of mean shift after MSD when changing the kernel bandwidth $b$ and the seed radius $r$.}
  \label{fig:compare-b}
\end{figure*}

We subsequently investigated the effects of changing the kernel
bandwidth $b$ (see Eq.~\ref{eq:kernel}) on images produced by MSD. While larger values of $b$
tend to push the tradeoff between precision and recall slightly in
favor of the former, the $F_1$ measure remains statistically
indistinguishable (using 95\% confidence intervals). Results are
reported in Figure~\ref{fig:compare-b}, where barplots are used for
better readability.

\subsection{Effects of the new variants for neural network architecture and training}
\begin{figure*}
  \centering
  \includegraphics[width=\textwidth]{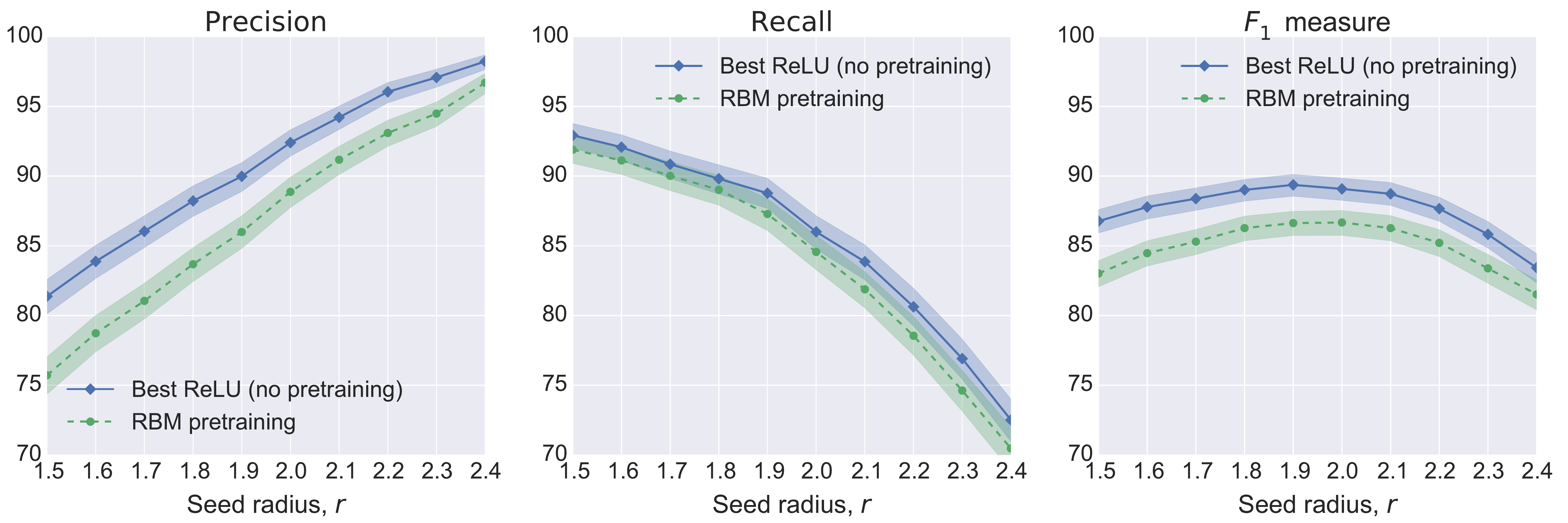}
  \caption{MSD: comparison between the former finely tuned RBM
    architecture~\cite{frasconi_large-scale_2014} and the new
    architecture based on ReLUs, columnar structure, and masked
    training.}
  \label{fig:compare-best-relu-vs-rbm}
\end{figure*}

Semantic deconvolution in~\cite{frasconi_large-scale_2014} was based
on an neural network architecture with pre-trained restricted Boltzmann
machines (RBM) in the first two layers, followed by fine tuning by
backpropagation. In this paper, besides the multiview extension, we
have introduced some architectural variants, namely the use of
rectified linear units (ReLU) trained with backpropagation from random
initial weights, the columnar architecture shown in
Figure~\ref{fig:MSD-architectures}(b), and the use of a masked
training (randomly zeroing one of the two input views). While
none of these three variants alone yields a significant performance
improvement, their combination does. We compared the two methods in
the MSD setting. Results in Figure~\ref{fig:compare-best-relu-vs-rbm}
show a slight improvement in recall and a significant improvement in
precision, for all values of $r$. The improvement in terms of $F_1$
measure is also significant (using 95\% confidence intervals).

\subsection{Mapping neuronal activity at the whole brain scale}
\label{sec:mapping}
\begin{figure*}
  \centering
  \includegraphics[width=\textwidth]{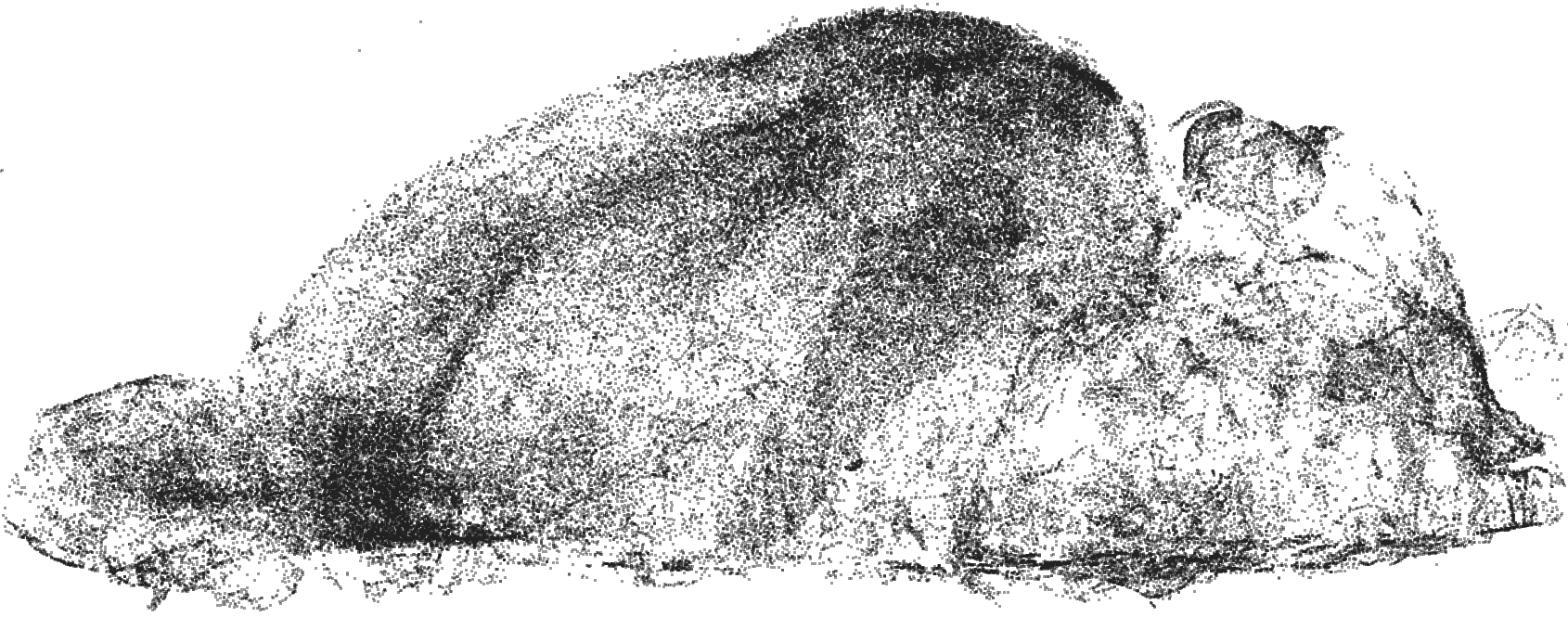}
  \caption{Point cloud showing a whole-brain activation map. $91,584$
    active cells have been detected by the software.}
  \label{fig:cloud}
\end{figure*}

We finally applied our method to the whole Arc-dVenus mouse brain
images. The registration block (see Figures~\ref{fig:pipeline}
and~\ref{fig:pipeline-registration}) ran in about $\SI{720}{\minute}$
using a cluster of eight quad-core Intel(R) Xeon(R) @2.40GHz
CPUs. Multiview semantic deconvolution required $\SI{800}{\minute}$
using two Tesla K40 GPUs. Finally, cell detection and fusion
required $\SI{40}{\minute}$ on the CPU cluster. Running time of course
depends on the number of non-black substacks, which are 4194 out of
$15,552$ in the present case.
After cell detection, we obtained 3622 non-empty substacks with a total of
$91,584$ detected cell bodies. Figure~\ref{fig:cloud} shows the whole
activation map obtained with our approach. Activated cells can be largely found in different layers of the cerebral
cortex, as well as in the hippocampus and the olfactory bulb. Fewer cells can also be observed in the cerebellar
cortex and in deeper brain areas.

\section{Conclusions}
We have presented a method for identifying cell bodies at the
whole-brain scale and applied it to the mapping of neuronal activity
in Arc-dVenus mouse. The cell identification approach used in this
paper extend previous work~\cite{frasconi_large-scale_2014} to handle
multiview images and some improvements to the neural network used for
semantic deconvolution have been presented (use of ReLUs and 
masked training). Our results indicate that a specially designed multiview
semantic deconvolution (MSD) module taking as input two adjacent views
simultaneously outperform simpler approach based on content-based
image fusion (IFI) or independent processing of the individual views
(SVIM). Additionally, we have shown that semantic deconvolution is
able to significantly increase performance and stability of the
results with respect to changes in the parameter values, thus enabling
whole-brain analysis without the need of tuning parameters locally to
handle the quality variability problem in CLSM images. Our best
performance ($F_1$ measure of 89.4) cannot be directly compared to the
performance ($F_1$ measure of 96.0) we previously attained on
GFP-labeled Purkinje cells in the cerebellum. First, resolution in the
present study is significantly lower ($\SI{4}{\micro\meter}$
vs. $\SI{1}{\micro\meter}$
in~\cite{frasconi_large-scale_2014}). Second, the use of higher laser power
to penetrate the entire brain led to saturation of brightest voxels close to laser
entrance point.
Third, Purkinje cells have a special spatial arrangement
into 3D folia which enabled the use of manifold modeling to filter
false positive detections and gain 3 points of $F_1$ measure. In the present study,
since Arc expression is not related to a specific cell type, no \emph{a priori}
information on spatial arrangement of soma can be exploited. Our current performance
is anyway better than that reported in previous work addressing IEGs mapping \cite{kim_mapping_2015}.

The approach presented in this Article can provide an important tool to
understand whole-brain dynamics in health and disease. To this aim, the soma
point clouds obtained need to be mapped on a standard reference atlas
to allow quantitative comparison between different subjects. Further, the throughput of the entire
experimental pipeline has to be expanded and all the steps, including specimen clearing, imaging, stitching, cell identification
and atlasing, needs to be standardized and better integrated.
When applied to a significant cohort of mice, in different behavioral tasks and with
distinct genetic backgrounds, the methods described here can provide a better understanding
of the principles that orchestrate
neuronal activity across the entire brain in health and disease.

\section*{Availability}
The python software used in this paper is included in the 1.1 release
of \texttt{bcfind}~\cite{frasconi_large-scale_2014} and employs
portions of \texttt{pylearn2}~\cite{goodfellow_pylearn2:_2013} and
\texttt{scikit-learn}~\cite{pedregosa_scikit-learn:_2011}. It is released in open source form
under GPL3 and is available from \url{https://github.com/paolo-f/bcfind}. 
The Arc-dVenus mouse brain images
are available at
\url{https://dataverse.harvard.edu/dataverse/arc_dvenus}.

\begin{acknowledgements}
  We would like to thank Nikita Rudinskiy e Bradley T. Hyman for
  providing us with a specimen of the animal that was used to develop
  and validate the methods described in this paper.
\end{acknowledgements}

\bibliographystyle{spmpsci}      
\bibliography{nibrain}   

\end{document}